%% file: main.tex
\documentclass{isprs} %
\usepackage{subfigure}
\usepackage{setspace}
\usepackage{geometry} %
\usepackage{epstopdf}
\usepackage{amsmath}
\usepackage[labelsep=period]{caption}  %
\usepackage[british]{babel} 
\usepackage[hang]{footmisc}

\usepackage{natbib}

\usepackage{url}

\usepackage{makecell}
\usepackage{xcolor}
\usepackage{booktabs}
\usepackage{tikz}
\usepackage{multirow}
\usepackage{url}
\graphicspath{{figures/}}

\begin{document}

\title{DETECTION OF DEGRADED ACACIA TREE SPECIES USING DEEP NEURAL NETWORKS ON UAV DRONE IMAGERY}
\date{}

 \author{
  Anne Achieng Osio\textsuperscript{a,}\thanks{Corresponding author. Email: \protect\url{osio@univ-ubs.fr, ansiyo22@gmail.com}} $\qquad$ Hoàng-Ân Lê\textsuperscript{b}$\qquad$   Samson Ayugi\textsuperscript{a}$\qquad$  Fred Onyango\textsuperscript{a}$\qquad$  Peter Odwe\textsuperscript{a}$\qquad$  Sébastien  Lefèvre\textsuperscript{b}}
 
 \address
 {
 	\textsuperscript{a }The Technical University of Kenya (TUK),  Faculty of Engineering \& Built Environment, Nairobi, Kenya \\
 	\textsuperscript{b }IRISA, Université Bretagne Sud (UBS), Vannes, France
 }

\commission{III, }{III} %
\workinggroup{III/10} %
\icwg{}   %

\abstract{
Deep-learning-based image classification and object detection has been applied successfully to tree monitoring. However, studies of tree crowns and fallen trees, especially on flood inundated areas, remain largely unexplored. Detection of degraded tree trunks on natural environments such as water, mudflats, and natural vegetated areas is challenging due to the mixed colour image backgrounds. In this paper, Unmanned Aerial Vehicles (UAVs), or drones, with embedded RGB cameras were used to capture the fallen Acacia Xanthophloea trees from six designated plots around Lake Nakuru, Kenya. Motivated by the need to detect fallen trees around the lake, two well-established deep neural networks, i.e. Faster Region-based Convolution Neural Network (Faster R-CNN) and Retina-Net were used for fallen tree detection. A total of 7,590 annotations of three classes on 256$\times$256 image patches were used for this study. Experimental results show the relevance of deep learning in this context, with Retina-Net model achieving 38.9\% precision and 57.9\% recall.

}


\maketitle

\sloppy

\section{Introduction}\label{INTRODUCTION}

Forest detection has been embraced in many studies using different types of remotely sensed datasets, captured from different aerial and satellite sensor platforms. Assessment of fallen trees is an important step to characterize forest health. According to~\citet{torres2021role}, most of the  studies have been applied mainly in America and Europe with the most frequently used data being multipectral imagery from Landsat sensors~\citep{she2015comparison}. The recently launched  Sentinel-2 have shown less capability in detecting individual vegetation species~\citep{nzimande2021mapping}. UAV remotely-sensed images are ideal for forest health assessment since they provide optical imagery with high geometric spatial resolution (10-40cm)~\citep{naik2021prediction}. Conversely, coarse ground resolution satellite imagery from optical sensors does not allow to capture the geometric structure of fallen trees~\citep{naik2021prediction}, and hence remains mainly used for classification at local and regional scales~\citep{gorelick2017google}. Other studies have also shown that incorporating LiDAR data with multispectral imagery improved the prediction of tree height estimation and canopy detection models within natural forests~\citep{manzanera2016fusion}. 

Despite the reliability of LiDAR data, its means of acquisition  remains more  expensive than that of UAV based data~\citep{ampatzidis2019citrus}. Moreover, models based on LiDAR data (especially in relation to above ground biomass or fallen trees) highly depends on the type of forest under study and the level of tree degradation on site~\citep{galidaki2017vegetation}. Synthetic Aperture Radar (SAR) data with longer wavelengths and cross-polarization capabilities have also been used in fallen tree studies. Models created using SAR data  produce uncertainty and variance in modelling accuracy of above ground biomass~\citep{dalponte2018predicting,naik2021prediction}. On the contrary,~\citet{osio2021object} confirmed that the use of SAR-C channels captured in Single Look Complex mode in conjunction with machine learning models and object-oriented approach yielded the best results with an Overall Accuracy (OA) of 98.1\% and a Kappa of 97.0\%, hence improving the above ground biomass classification on the Acacia xanthophloea strands around Lake Nakuru, Kenya. Despite achieving results at local scale, such a model was not suitable to capture individual degraded Acacia xanthophloea target trees that are fallen around the lake. In recent times, the deep learning paradigm with models tailored at image classification and detection has become a standard methodology in remote sensing studies. The main advantage provided by deep learning over classical machine learning approaches is that models created using deep learning can learn and extract information directly from input data, not requiring a costly feature engineering step. These models can then be used to detect and predict similar features on the entire scene under investigation.

\begin{figure*}[t]
\centering
\includegraphics[width=\linewidth]{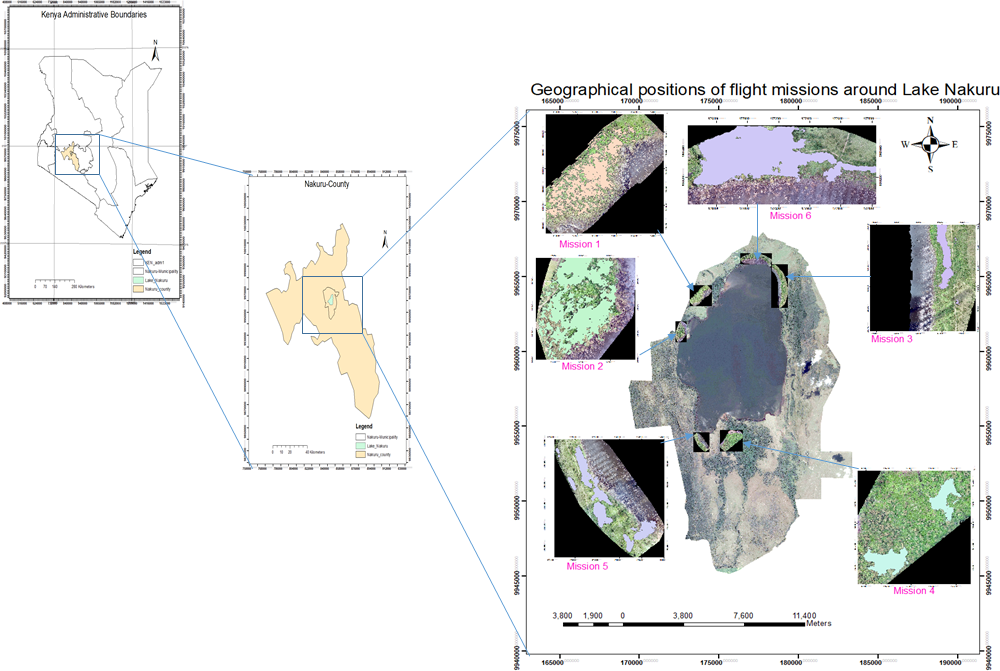}
	\caption{The geographic position of UAV flight missions around Lake Nakuru, projected at WGS84, UTM Zone 37S.}
	\label{fig:flight}
\end{figure*}

Deep learning has been used in many studies in recent times including automatic extraction of ice-wedge polygons using Mask R-CNN framework on both high-resolution imagery and UAV, producing F1 Scores of 72\% and 70\% respectively~\citep{zhang2020transferability}.~\citet{santos2019assessment}  made a comparison of three different deep learning frameworks, namely YOLOv3, Faster Region Based Convolutional Neural Networks (Faster R-CNN), and RetinaNet to assess a time series of RGB images in the context of tree crown detection achieving an overall average precision (AP) of 92\%. Other studies have used object-based image analysis approach to detect coarse wood debris (CWD) from unmanned aerial systems in conjunction with LiDAR point clouds~\citep{thiel2020uas}. The authors reported an overall average precision (mAP) of 85\% and a recall of 69.2\%. However, they pointed out that the results achieved using very high resolution imagery and their line detection algorithm over CWD areas could be improved using deep learning approaches~\citep{jiang2019dead}. Further concerns by the authors involved controversy about the application of deep learning frameworks on tree species, pointing out the challenge in model transferabilty onto similar scene.

In this paper, we evaluate off-the-shelf deep architectures in detecting fallen trees of the Acacia Xanthophloea trees around Lake Nakuru National Park, Kenya.  The wetland was designated amongst wetlands of international importance~\citep{odada2004experiences} as Ramser site number 476 on 5th June, 1990. Kenya, being a signatory to the Conference of Parties (COP) based on the Ramser convention in 1971~\citep{davidson2019review} is required to actively conserve and make wise use of Lake Nakuru wetlands in a sustainable manner.

It is important to quantify dead woody Acacia xanthophloea since, according to previous studies their presence improves  biological diversity within live forests such as the introduction of mosses and lichens which attracts migratory birds in the National Park \citep{vareschi1985ecology,harmon1986ecology,norden2008partial}. The downed Acacia xanthophloea was not caused by any climatic factors but rather by the increased volume of water due to sedimentation \citep{iradukunda2020sedimentation} in Lake Nakuru which overflown its bank \citep{osio2018obia}, hence weakening the riparian trees from their roots, causing them to fall. 
Coarse fallen woody debris in water-bodies \emph{i.e.} lakes and streams are known to increase channel complexity, which contributes to the improvement of habitat quality hence increasing nutrient retention inside  stream systems \citep{cowden2002study,swanson1992promoting}.
Previous research by \citet{bisson1992best} reported reduction in habitat quality in streams that underwent traditional clearance of the deadwood in their stream systems.

UAV flight  missions around the lake revealed massive destruction of the trees, especially inside the water body and on the mudflats. Therefore, the purpose of this study was to provide a state-of-the-art model based on the detection and classification of fallen trees around the lake, hence enabling the wildlife and forest conservation managers to make informed decision on the fallen Acacia xanthophloea trees.
More precisely, our main goal is to evaluate the performance of well-established deep neural networks over UAV-based fallen tree datasets.To the best of our knowledge, there are  no known studies that have been carried out on the detection of fallen Acacia xanthophloea around Lake Nakuru, using UAV/RGB in conjunction with deep learning approaches.

\begin{figure*}[htb]
\centering
\includegraphics[width=.32\textwidth]{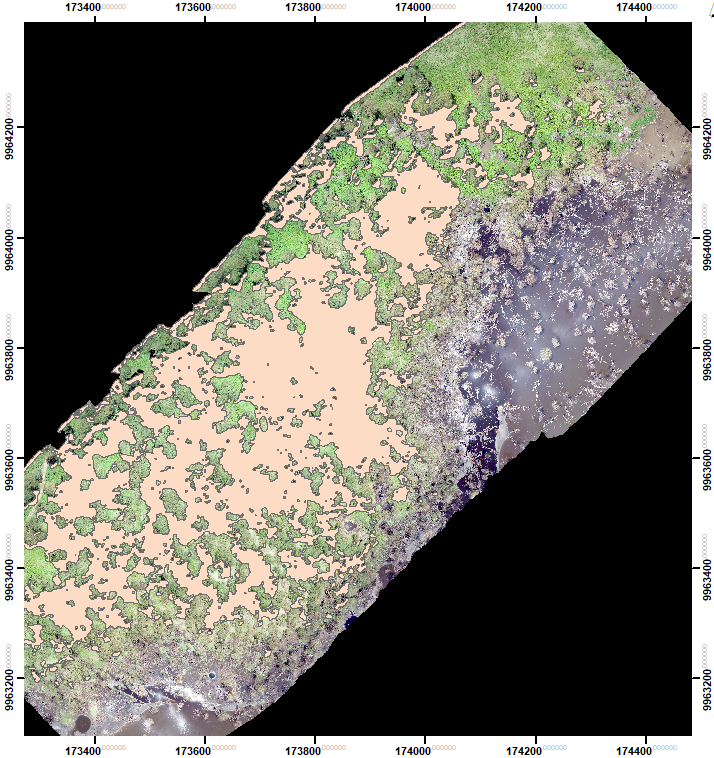}\hfill
\includegraphics[width=.32\textwidth]{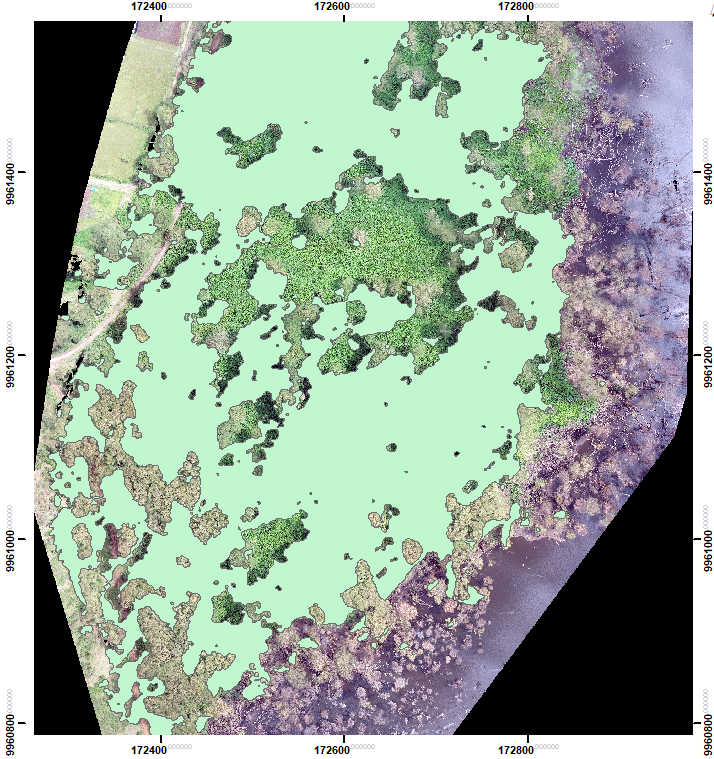}\hfill
\includegraphics[width=.32\textwidth]{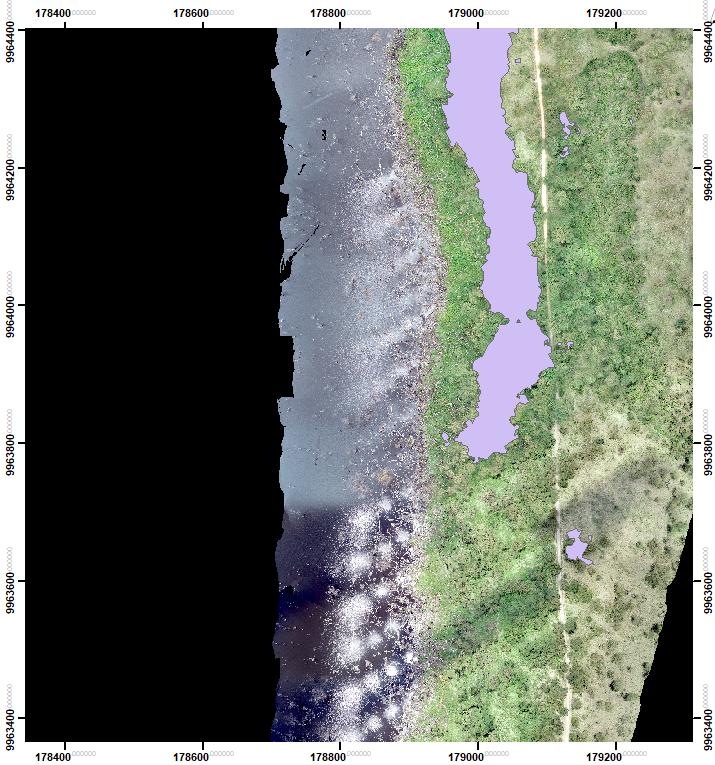}\\~\\
\includegraphics[width=.32\textwidth]{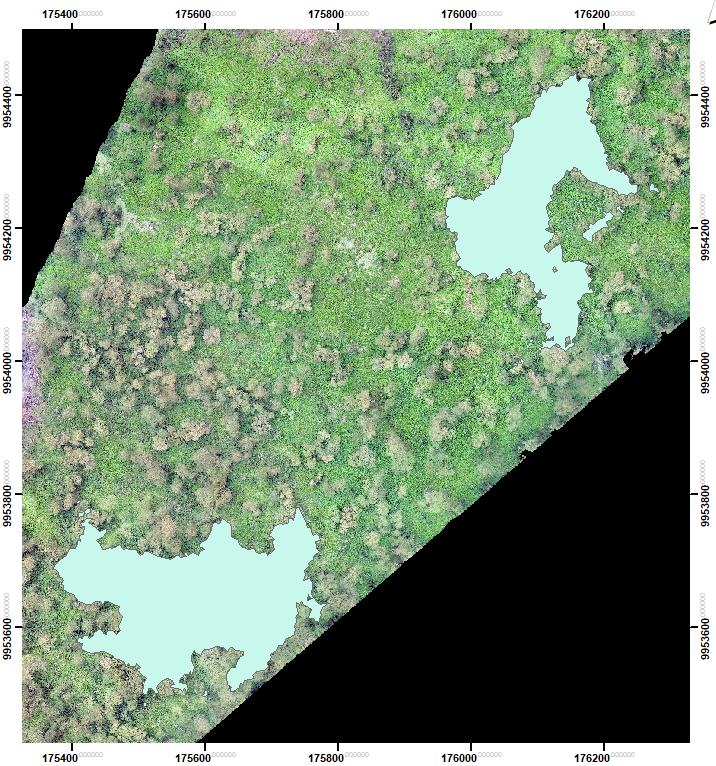}\hfill
\includegraphics[width=.32\textwidth]{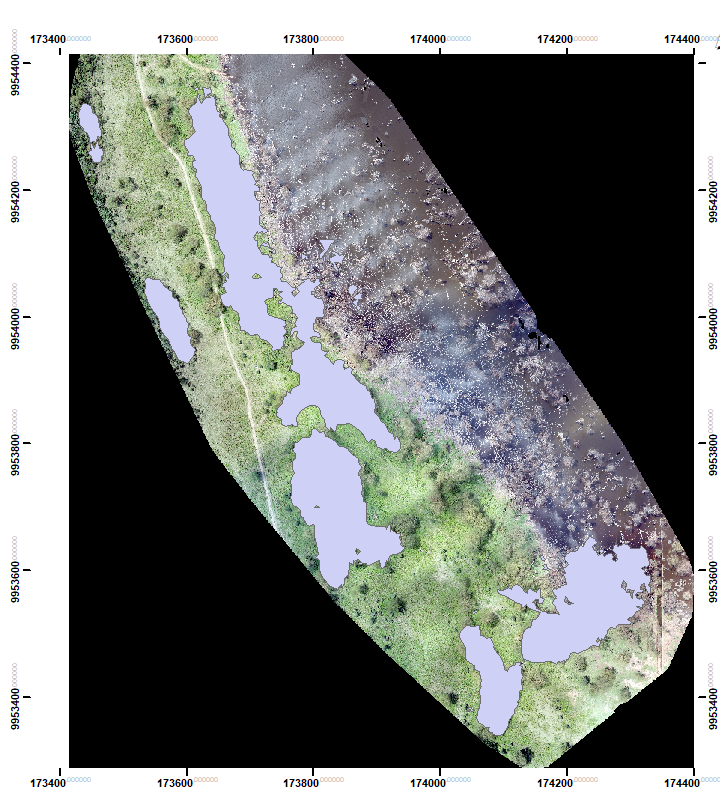}\hfill
\includegraphics[width=.32\textwidth]{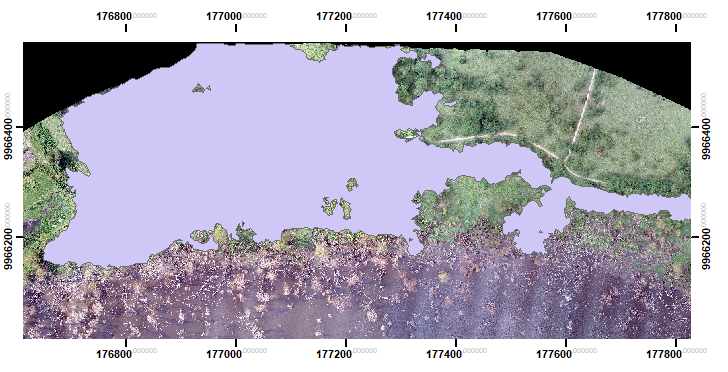}
	\caption{The six flight missions captured from different sites around the Lake using DJI Phantom 4, SDK Drone. The overlay vector polygons were derived from QGIS-based forest detection plugin, Mapflow.ai representing the detected trees on each imagery.} \label{fig:map1}
\end{figure*}

\section{Materials and Methods}\label{sec:Materials and Methods}

\subsection{Study Site}\label{sec:study_site}

Lake Nakuru National park is located in Nakuru County (see Fig.~\ref{fig:flight}), approximately 170 km away  from Nairobi. The Park is situated geographically at  Latitudes 0$^{\circ}$18'S and 0$^{\circ}$27'S and Longitude 36$^{\circ}$1.5'E and 39$^{\circ}$9.25'E within the Kenyan Rift valley \citep{mubea2012monitoring}.  Before the recent flooding, its bottom was initially about 1,756 m above sea level while the surface of the water was at 1,758.5m above sea level. The altitude ranges from 1,760-2,080m above sea level~\citep{iradukunda2020sedimentation}. Mean annual rainfall ranges between 876mm and 1,050mm and has an inherent bi-modal pattern \citep{odada2004experiences}. The long rains start in March and end in June while the short rains occur between October and December. Mean daily minimum and maximum temperatures fluctuate between 8.2$^{\circ}$C and 25.6$^{\circ}$C~\citep{ng2010distribution}. Lake Nakuru has no outlets and hence evaporation is the only factor that accounts for water loss. Four seasonal rivers feed the lake,~\emph{i.e.} Lamurdiak, Makalia, Enderit and Enjoro. Acacia xanthophloea tree  patches have been in existence on the Northern, Southern, Eastern and Western side of the lake for decades. The soils on the shores of this lake are volcanic and shallow in nature. Underneath the Acacia savanna were the open grasslands thriving on soils and ashes that were well-drained, friable to sandy clay loams (see Fig.~\ref{fig:map1}). Recent studies have shown that the health of Acacia xanthophloea trees  have been degrading since the year 2010 due to the persistent flooding around the lake~\citep{osio2020spatial,osio2021object}. 

\begin{figure*}[t]
\centering
	\includegraphics[width=\linewidth]{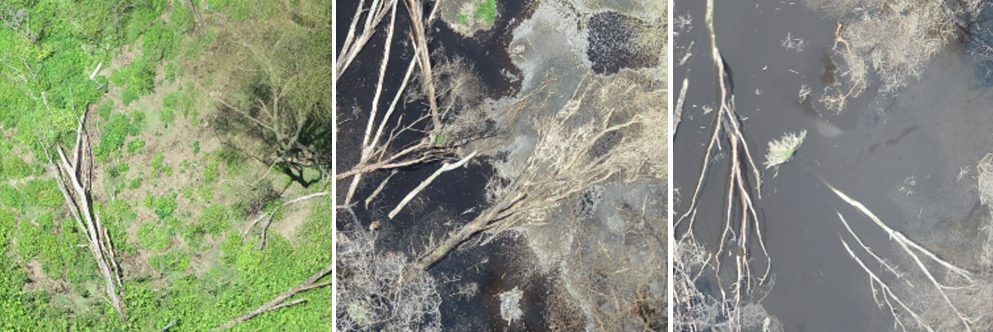}
	\caption{The three sample classes and numbers of annotations in the dataset: from left to right, dead trees on land (2,514 boxes), dead trees on mudflat (2,077 boxes), and dead trees on water (2,999 boxes).}
	\label{fig:study}
\end{figure*}

\subsection{Data Capture}\label{sec:Data capture}

Unmanned Aerial Vehicles (UAV) drone imageries were captured across six sites around the lake in the early September 2021 for 5 consecutive days, see Fig.~\ref{fig:map1}. The UAV drone used for the ground surveys was a DJI Phantom 4 RTK SDK, all its specification are reported in Tab.~\ref{tab:drone}. Images with a mean Ground Sampling distance (GSD) of 4.84cm and covering approximately 6.1 square kilometers were captured across six designated sites around the Lake Nakuru National Park. The surveyed areas were mainly near the lake shorelines where a large number of degradation were observed during the flight planning stage. A total of 9,056 training image patches were generated from the six images (see Figure~\ref{fig:study}) with 7,590 tree samples annotated from the training sets. The fallen trees are classified into 3 classes according to their background, namely Water (W), Land (L), and Mudflat (M) as shown in Fig.~\ref{fig:study}.

\subsection{General Workflow}\label{sec:General workflow}

The study consisted of 4 steps. On the first step, images are acquired from a UAV/RGB drone platform. Then, patches of non-overlapping 256-pixel squares are extracted from the acquired images and randomly sampled to training and test set. The fallen Acacia xanthophloea  trees from the image patches are annotated with bounding boxes using the LabelImg\footnote{\url{https://github.com/tzutalin/labelImg}} tool. Finally, the training image patches with their annotations are used to train a deep convolutional neural network (CNN) while the mutually exclusive test set is used to report the performance of the network on unseen data.

\paragraph{Deep learning for object detection}\label{sec: OBJECT DETECT)}

Object detection is made of 2 subtasks, (1) localizing an object of interest
in an image with a bounding box and (2) categorizing the box with the correct class.
The tasks could be performed in a 2-step process, with the bounding boxes first
proposed then classified, or all at once, leading to the so-called 1-stage detectors. The exemplar model for each type of
architecture include Faster Region based Convolutional Neural Networks (Faster RCNN)~\citep{ren2015faster} and RetinaNet~\citep{retinanet},
respectively. In this study, we evaluate these two models for detecting fallen trees
around the Lake Nakuru.

\paragraph{Faster RCNN}
The Region-based CNN (RCNN) laid the groundwork for deep-learning-based object
detection~\citep{girshick2014rich} with bounding boxes proposed by Selective
Search~\citep{selective_search}. Fast R-CNN~\citep{gkioxari2015contextual} speeds
up the process by introducing the region of interest (ROI), pooling layer and inputting the full image
instead of just the proposals to the deep network. Significant acceleration is
achieved by Faster RCNN when a sub-network, called Region Proposal Network (RPN),
is used to generate the proposal boxes in place of Selective Search and both 2
stages can be trained end-to-end.

The vanilla Faster RCNN architecture extracts low-level image features by passing
an input image through several convolutional blocks, called a backbone (sub-)network.
The features are then shared between both the region proposal network (RPN) and
region of interest (ROI) head. The Region Proposal Network (RPN) generates a number of proposal boxes, 2000
by default, from which the corresponding features are obtained and classified by the
ROI head (sub-)network. The classifier is a multi-layer perceptron (MLP).

\paragraph{RetinaNet}
RetinaNet~\citep{retinanet} is a single-stage architecture for object detection. It consists of (i) a classification sub-network which predicts the probability of an object occurrence at each spatial location for each annotated box and object class, (ii) a regression sub-network that regresses the offset for the bounding boxes from the annotated boxes for each ground-truth object, (iii) a bottom-up pathway which consists of the backbone network (ResNet) whose role is to calculate the feature maps at different scales, and (iv) a top-down pathway that  up-samples the spatially coarser feature maps from higher pyramid levels. Lateral connections are included to merge top-down layers and bottom-up layers with the same spatial size.
Specifically, the Feature Pyramid Network (FPN) is proposed for feature extraction through upsampling or downsampling approach and relies on the focal loss objective function.%

\subsection{Experimental setup}

In this study, we use the implementation of Faster RCNN and RetinaNet provided in the Detectron2 library as previously implemented by~\citet{wu2019detectron2}.%

In relation to this experiment, all the annotations from the three classes, namely dead tree on Land (L), Water (W) and Mudflat (M) were sampled into two parts consisting of training/validation and testing set. From each given class of the dataset, 80\% of the samples across all the missions were used for training/validation while 20\% remained for testing. This particular design was adopted to cater for imbalances within the datasets across the 6 missions (Fig.~\ref{fig:map1}). Missions 3 and 4 have particularly fewer annotations compared to the rest of the missions. %

\paragraph*{Metrics}
We follow the Common Objects in Context (COCO) challenge
for quantitative assessment, as demonstrated by~\citet{lin2014microsoft}. The challenge employs the standard definition of 
precision (P) and recall (R) based on the notion of true positive (TP), false positive (FP), and false negative (FN):
\begin{equation}
    \label{eq:prf1}
    \text{P}=\dfrac{\text{TP}}{\text{TP}+\text{FP}} \quad
    \text{R}=\dfrac{\text{TP}}{\text{TP}+\text{FN}} \quad
    \text{F1}=2\dfrac{\text{P}\times\text{R}}{\text{P}+\text{R}}
\end{equation}

The COCO metric differs in the definition of a \emph{positive} box prediction,
for which the intersection over union (IoU, or Jaccard index) of it with ground
truth boxes are computed. IoU measures the ratio between the overlapping area of
the two boxes divided by the area covered by their union. As such, the number of
positive boxes changes according to the IoU level: higher IoU threshold (max
of 1 or 100\%) results in fewer positive predictions and thus, lower measurement.

At an IoU threshold, the boxes with at least (or higher than) the given level
are considered positive, true or false depending on the predicted class, and
are used to compute  precision and recall.
In this paper, the precision are computed for each of pre-defined recall values
(101 values from 0 to 1 with step of 0.01), which are used to plot the
precision-recall (PR) curve. We also report the average precision (AP),
average recall (AR), and subsequently F1-score using Eq.~\ref{eq:prf1} for 2 IoU
levels, 0.50 and 0.75.
The average precision is taken across all recall levels and equal to the area under
the PR-curve:
    \begin{equation}
        \label{eq:ap}
        \sum\limits_{j=1}^N p(k)\Delta r(k),
    \end{equation}
where $N$ is the total number of images in the collection, $p(k)$ is the precision
at a cutoff of $k$ images, and $\Delta r(k)$ is the change in recall that happened
between cutoff $k-1$ and cutoff $k$.

\begin{figure}[p]
\begin{center}
\includegraphics[width=\columnwidth]{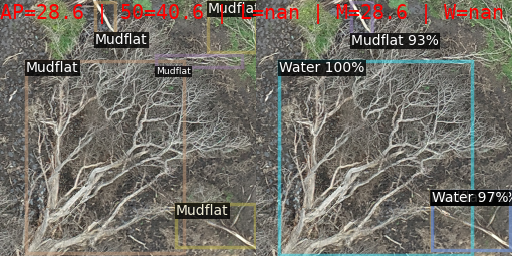}\\[.25em]
\includegraphics[width=\columnwidth]{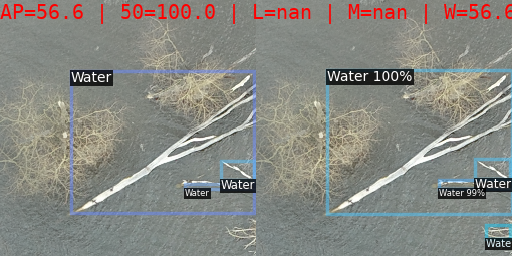}\\[.25em]
\includegraphics[width=\columnwidth]{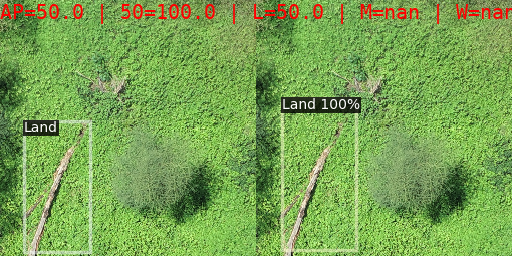}\\[.25em]
\includegraphics[width=\columnwidth]{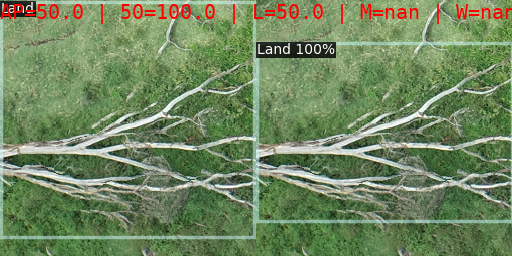}
	\caption{Sample results from different scenes using RetinaNet, with average precision (AP) averaging over all IoU levels in [0.5, 0.95], step size 0.05 and (50 being precision at $\text{IoU}\geq 0.5$) on three classes: Dead Tree in Water (W), on Land (L), and on Mudflat (M). Ground truths are shown on the left and predicted bounding boxes on the right. The confidence levels (\%) are shown on the bounding boxes.}
	\label{fig:retina}
\end{center}
\end{figure}

\begin{figure}[p]
\begin{center}
\includegraphics[width=\columnwidth]{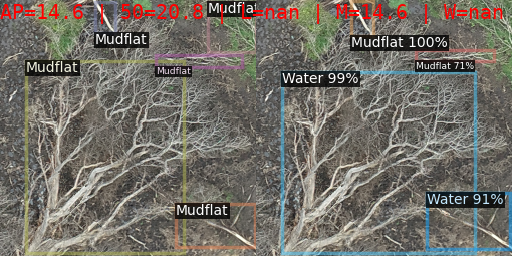}\\[.25em]
\includegraphics[width=\columnwidth]{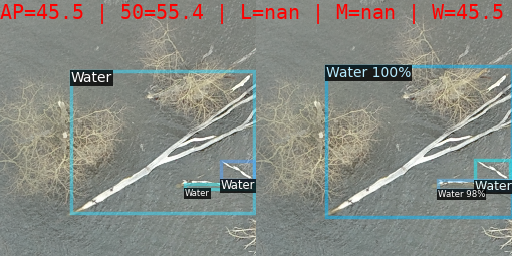}\\[.25em]
\includegraphics[width=\columnwidth]{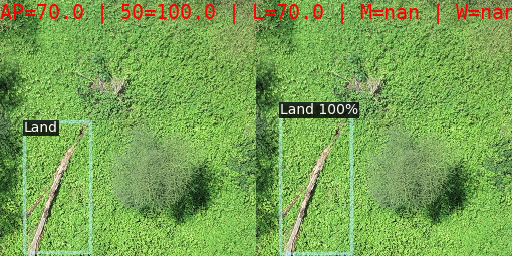}\\[.25em]
\includegraphics[width=\columnwidth]{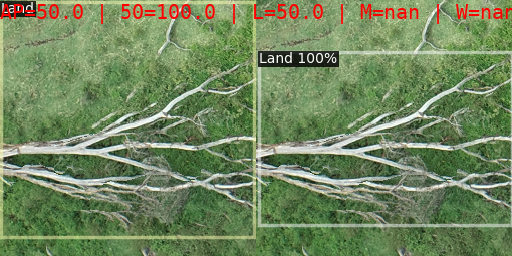}
	\caption{Sample results from different scenes using Faster-RCNN, with average precision (AP) averaging over all IoU levels in [0.5, 0.95], step size 0.05 and (50 being precision at $\text{IoU}\geq 0.5$) on three classes: Dead Tree in Water (W), on Land (L), and on Mudflat (M). Ground truths are shown on the left and predicted bounding boxes on the right. The confidence levels (\%) are shown on the bounding boxes.}
	\label{fig:rcnn}
\end{center}
\end{figure}

\section{Results}\label{sec:RESULTS AND DISCUSSIONS}

\begin{figure*}[t]
    \centering
    \def\svgwidth{.49\textwidth}
    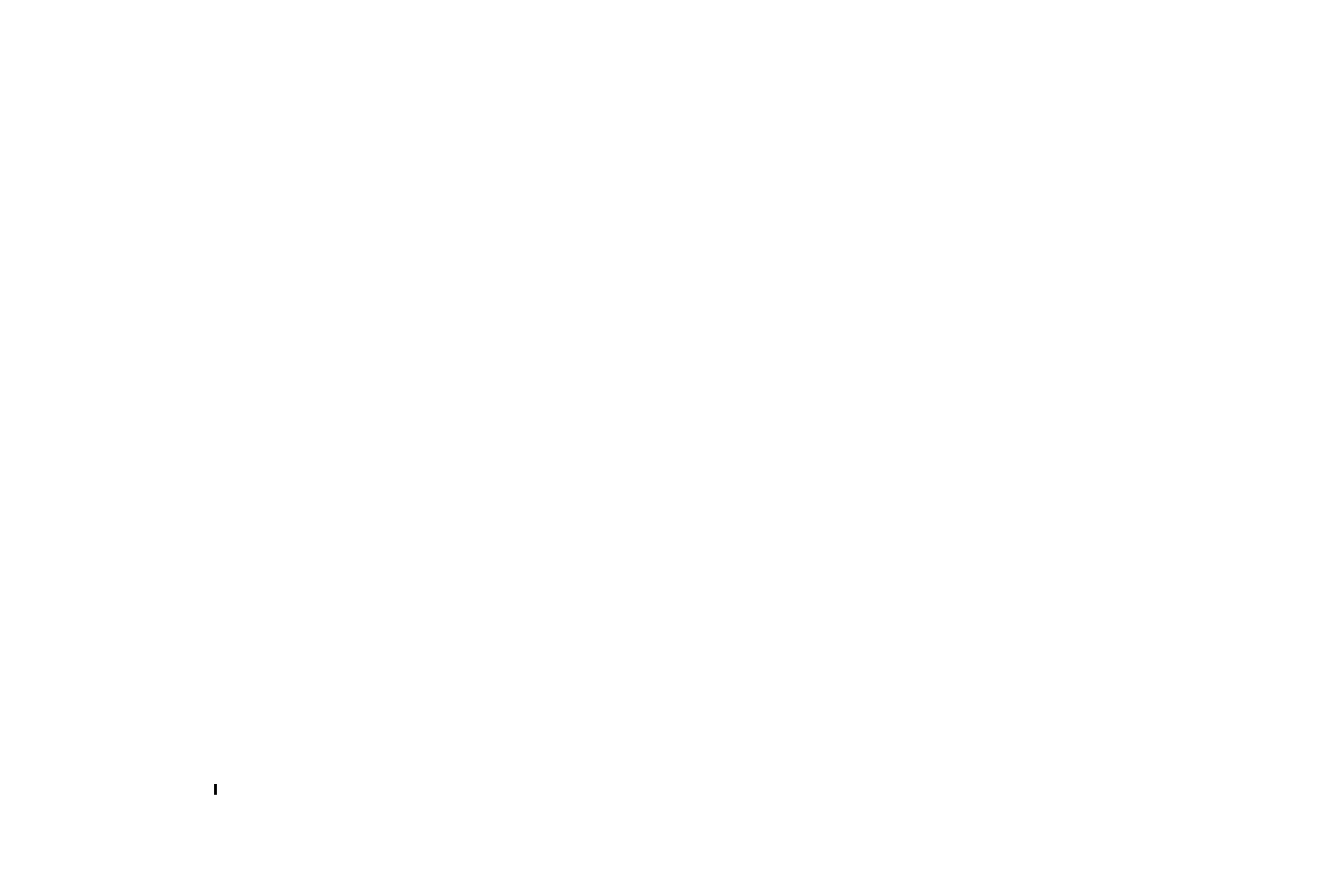
    \def\svgwidth{.49\textwidth}
    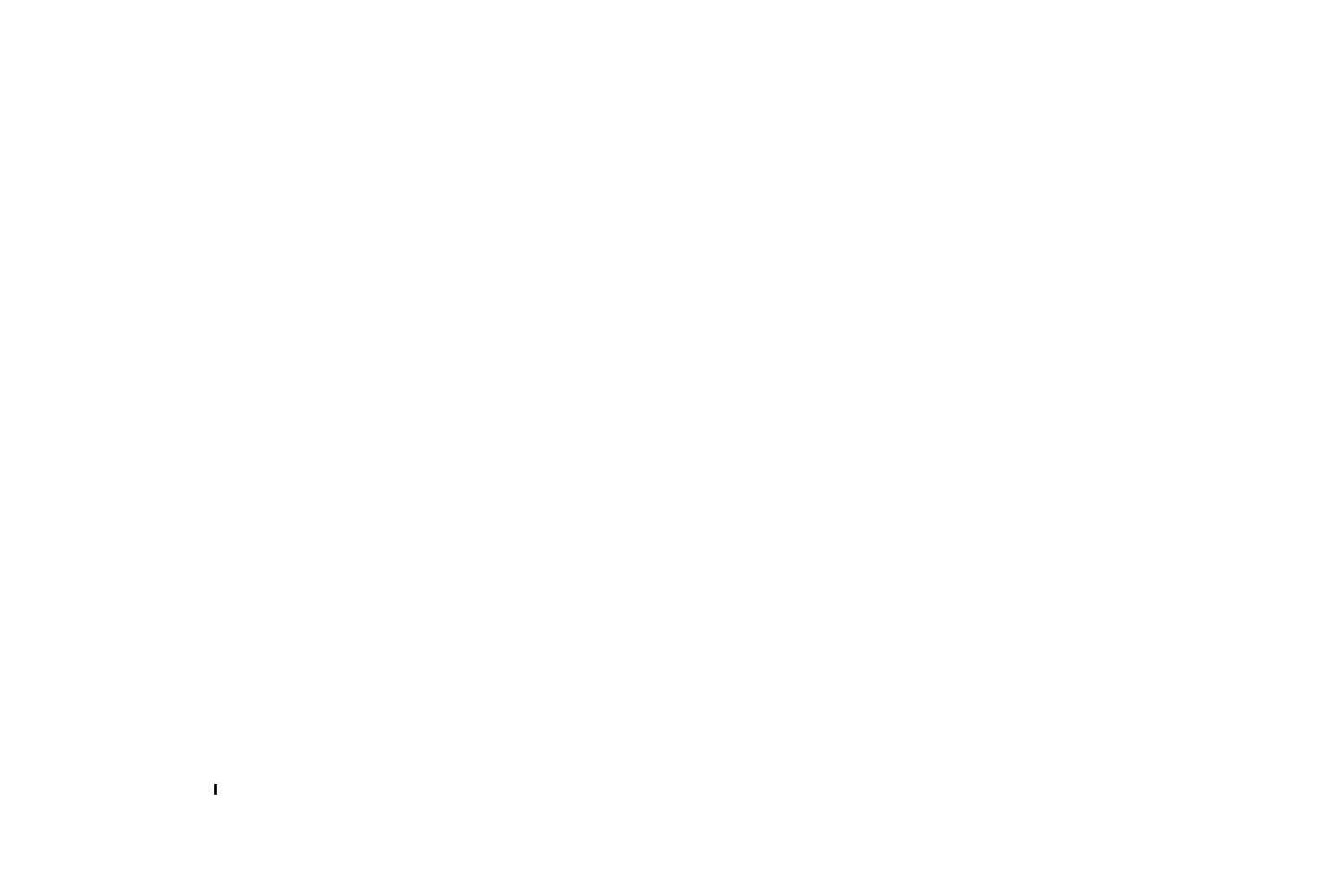
    \caption{Precision-Recall curves for  RetinaNet and two IoU thresholds.}
    \label{fig:retina_pr}
\end{figure*}

\begin{figure*}[t]
    \centering
    \def\svgwidth{.49\textwidth}
    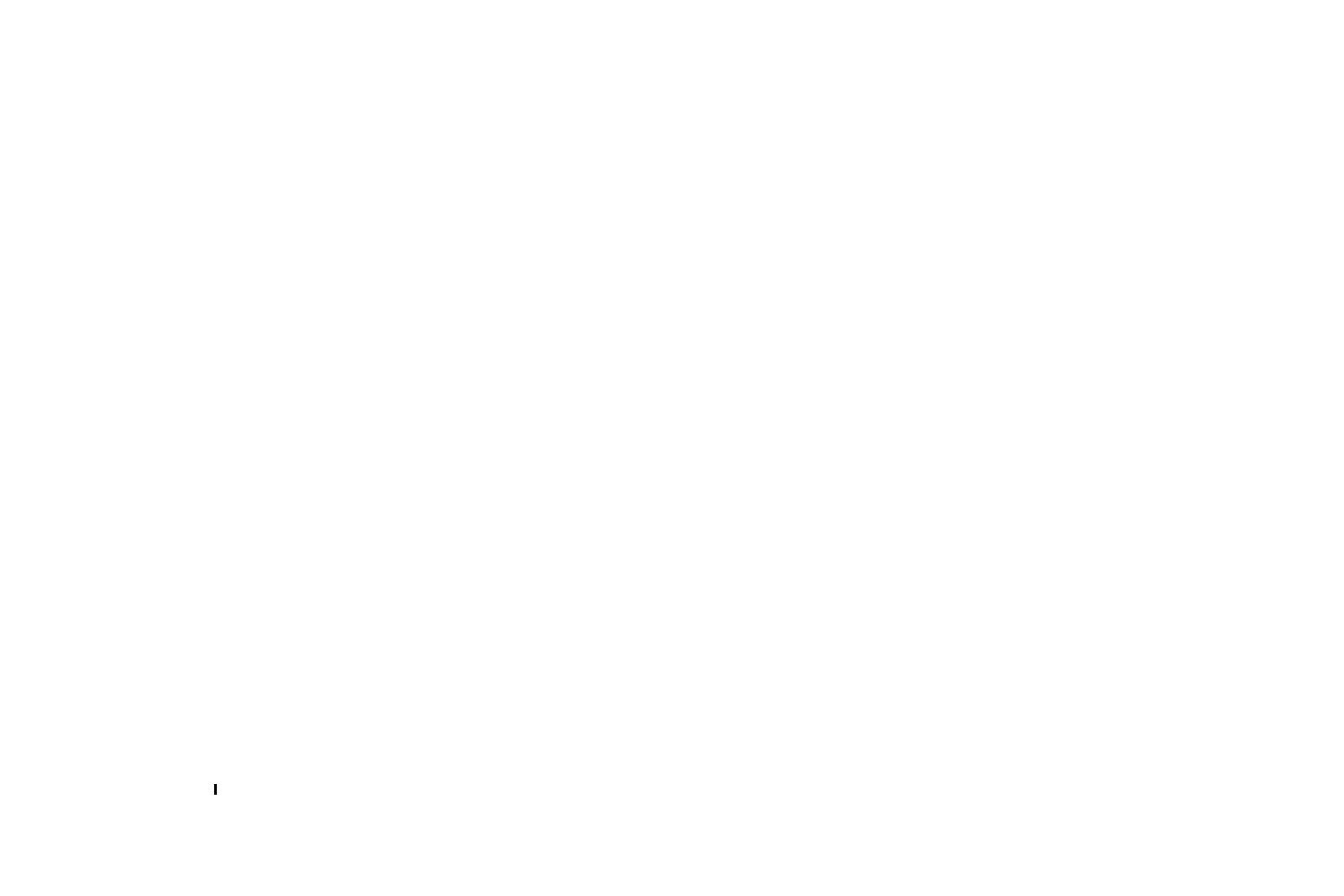
    \def\svgwidth{.49\textwidth}
    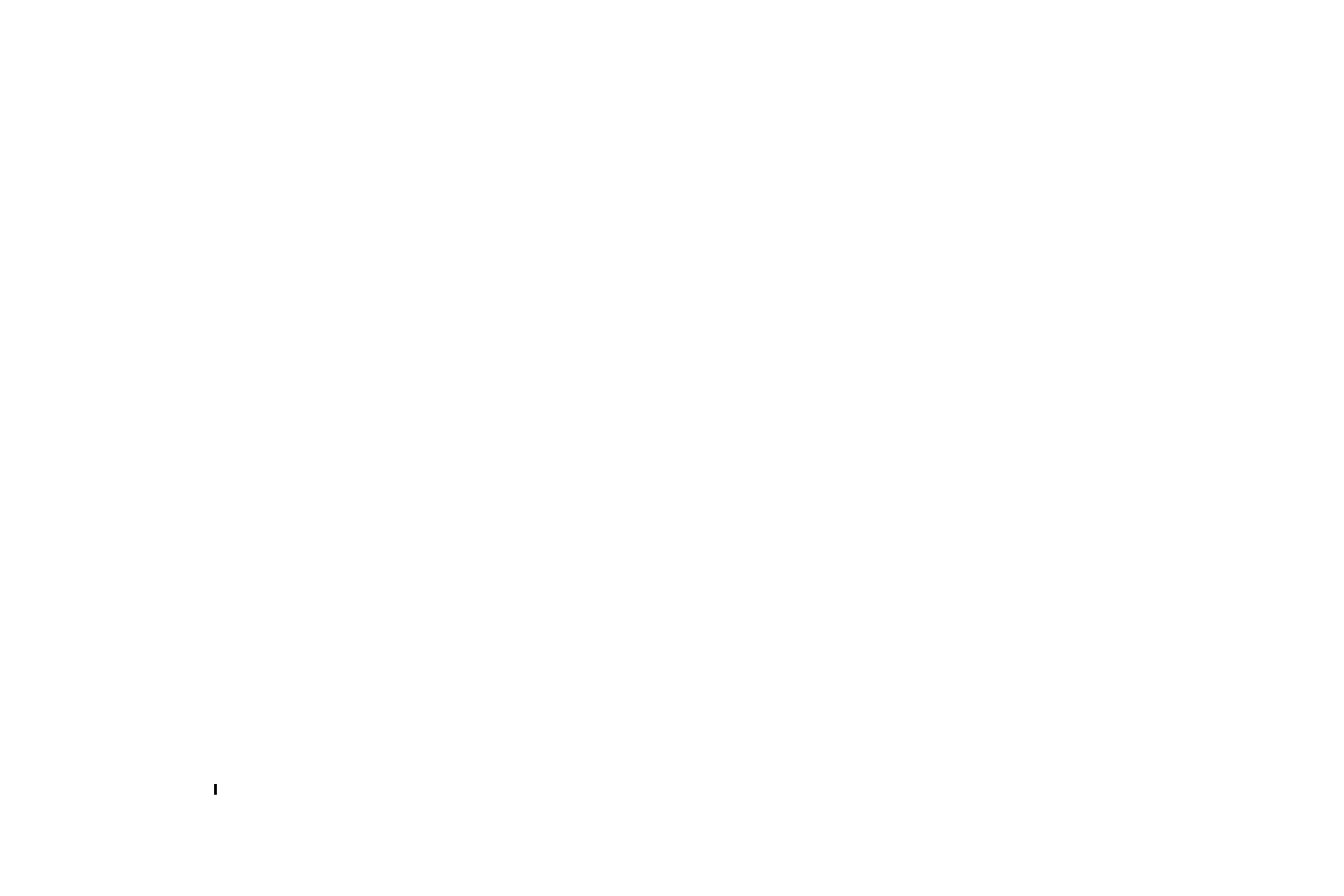
    \caption{Precision-Recall curves for  Faster RCNN and two IoU thresholds.}
    \label{fig:faster_pr}
\end{figure*}

We provide in Fig.~\ref{fig:retina_pr} and Fig.~\ref{fig:faster_pr} the Precision-Recall curves for the two deep neural networks considered in our study. When comparing  RetinaNet and Faster RCNN, one can observe that the former delivers higher average AP results (38.85\%) than the latter (37.21\%) considering an $IoU$ threshold of $0.5$. Nevertheless, the difference remains within a small margin.
In terms of Average Precision per class, RetinaNet performs best for both classes ``Land'' and ``Water'' (with improvement of 2-4\%) but worse for class ``Mudflat'' (degradation of 2\%). 

Considering a higher $IoU$ threshold of $0.75$, the methods reach lower APs. Again, RetinaNet performs better than Faster RCNN (also with a small margin). A classwise analysis leads to somehow different conclusions, with RetinaNet achieving better results for classes ``Mudflat'' and ``Water'', Faster RCNN for ``Land''. The margin being very small, one should take these results with caution.

We then show in Fig.~\ref{fig:retina} and Fig.~\ref{fig:rcnn} some visualization results achieved by RetinaNet and Faster RCNN per each given class. All classes were predicted with a high confidence level as shown at the edge of the bounding boxes. Illustrations also include the average Precision (AP) reported per class and mean Average Precision (mAP) on all the bounding boxes of a specific class (shown in red values).

\section{Discussion}
The model that achieves the best performances (Precision and Recall) was the RetinaNet with a setting $\text{IoU}\geq 0.5$, as shown in Tab.~\ref{tab:retina_results}. These results corroborates with recent studies carried out  by~\citet{santos2019assessment} where RetinaNet outperformed two other variants, namely YoloV3 and Faster-RCNN. Similar studies by~\citet{alon2019tree} involving tree crown detection with UAV orthophotos  and LiDAR point clouds had promising results with RetinaNet as well. 

The overall Average Precision (AP) reported with RetinaNet was 43.85\%. 
Usually AP or Area Under the Curve are normally used on imbalanced datasets. 
Datasets with low true positive rates (TPR) and high false positive (FP) and false negative (FN) rates usually produce low Precision and  Recall. 
In the case of our UAV dataset, annotations were carried out on the fallen trees which exhibited different backgrounds, \emph{i.e.} mudflat, water and land. Class ``water'' had the highest detection rates with an $\text{AP}\doteq 43.85\%$  due to uniformity of their background, while class  ``mudflat'' shows the opposite behavior due to the background non-uniformity.

It is well-known that a deep network's performance is impacted by the number of annotations and image patches derived from the UAV dataset, as already shown in recent studies by~\citet{hagele2020resolving}. Although a large number of annotations are employed in this study (9,056 patches with 7,590 annotations),
the performance is incomparable with tree crown detection using RetinaNet~\citep{santos2019assessment} which reportedly achieves AP of 92.64\% on 392 image patches.
We observed that two different areas might come with very different visual features and landscapes, hence problems with varying difficulty to tackle. More precisely, the tree crown detection problem is much simpler (thus leading to higher detection rates) due to the tree crown uniformity (shape factor) than our use case, where we had to deal with coarse wood debris (tree biomass) which come in different shapes and backgrounds, hence the creation of noise within networks.

We observed that the complexity of the proposed approaches which  are based on both networks were  trained on a cluster node with 
2 CPU x 20 E5-2687W v3 @ 3.10GHz, 396G RAM, and 1 NVIDIA GeForce GTX Titan-X of 12.2GB VRAM, which are shared among cluster users.
The Faster RCNN with ResNet50 backbone and FPN contains 17,260,319 parameters while the RetinaNet model with the same backbone contains around 14,460,660 parameters.

\section{Conclusion}\label{sec:conclusions}
In this study, we dealt with the detection of fallen Acacia xanthophloea  trees, on which we evaluated two competitive CNN models, \emph{i.e.} Faster-RCNN and RetinaNet using images captured by UAV with RGB Cameras on board. The networks were trained and assessed using a dataset made of 9,056 image patches and 7,590 annotations on bounding boxes. RetinaNet achieved overall Precision of 38.9\% and Recall of 57.9\%.

These results indicate that RGB Cameras embedded on UAV in conjunction with deep neural networks could possibly lead to the development of operational tools for the detection of fallen Acacia xanthophloea trees on different environmental backgrounds. This could also help in carrying out demographic surveys on fallen Acacia xanthophloea trees around the Park and in similar environments. Fallen Acacia xanthophloea tree demography could help ecologists and conservationists in quantifying the magnitude of Acacia tree degradation around Lake Nakuru. This model can be applied on other areas with similar characteristics such as fallen trees along the riparian reserve of other Rift valley lakes. Future studies should look into combining UAV-based images and their point clouds for tree detection and classification using deep networks.

\begin{table}[t]
	\centering
	    \setlength{\tabcolsep}{4pt}
		\begin{tabular}{@{}ll@{}}\hline
		    \toprule
			UAV Features & DJI Phantom 4 RTK\\
			\midrule
			 Frequencies used&GPS:L1/L2; GLONASS:L1/L2\\
			 Positional Accuracy& H: 1.5cm; V:1cm; Both +1ppm (RMS)\\
			 Image Sensor& CMOS 1"\\
			 Max resolution & 4864$\times$3648 (4:3); 5472$\times$3648 (3:2)\\
			 Field of view&84$^{\circ}$\\
			 Mechanical Shutter& 8-1/2000s\\
			 Data format& Photo (JPEG), Video (MOV)\\
		    \bottomrule
		\end{tabular}
	\caption{Da-Jiang Innovations Science \& Technology Co. Ltd (DJI) Real Time Kinematic (RTK) of Phantom 4 Specifications according to~\protect\citep{phantom2018rtk}.}
	\label{tab:drone}
\end{table}

\begin{table}[t]
    \centering
    \setlength{\tabcolsep}{3pt}
    \begin{tabular}{@{}ccccccc@{}}
         \toprule
         \multirow{2}{*}{IoU} & \multicolumn{3}{c}{RetinaNet} &  \multicolumn{3}{c}{Faster RCNN} \\
         \cmidrule(r){2-4}\cmidrule(l){5-7}
         & Precision & Recall & F1 Score &  Precision & Recall & F1 Score \\
         \midrule
         0.50 & 38.9\%  & 57.9\% & 46.5\%  & 37.2\% & 53.2\% & 43.8\% \\
         0.75 & 14.1\%  & 29.0\% &19.0\%  & 14.0\% & 27.7\% & 18.8\% \\
         \bottomrule
    \end{tabular}
    \caption{Quantitative results obtained with the two deep neural networks and two IoU thresholds.}
    \label{tab:retina_results}
\end{table}

 \section{Acknowledgements}\label{ACKNOWLEDGEMENTS}
                       
 The authors acknowledge:  Kenya National Research Fund (K-NRF) and Campus France through Pamoja PHC, Kenya National Council for  Science Technology \& Innovation (K-NACOSTI) and Kenya Wildlife Services for providing permit to enable  Drone Surveys in Lake Nakuru.

{%
    \renewcommand{\bibsection}{\section*{References}}
	\begin{spacing}{0.9}%
		\bibliography{main} %
	\end{spacing}
}

\end{document}

%% file: figures/retina_50.pdf_tex
\begingroup%
  \makeatletter%
  \providecommand\color[2][]{%
    \errmessage{(Inkscape) Color is used for the text in Inkscape, but the package 'color.sty' is not loaded}%
    \renewcommand\color[2][]{}%
  }%
  \providecommand\transparent[1]{%
    \errmessage{(Inkscape) Transparency is used (non-zero) for the text in Inkscape, but the package 'transparent.sty' is not loaded}%
    \renewcommand\transparent[1]{}%
  }%
  \providecommand\rotatebox[2]{#2}%
  \newcommand*\fsize{\dimexpr\f@size pt\relax}%
  \newcommand*\lineheight[1]{\fontsize{\fsize}{#1\fsize}\selectfont}%
  \ifx\svgwidth\undefined%
    \setlength{\unitlength}{432bp}%
    \ifx\svgscale\undefined%
      \relax%
    \else%
      \setlength{\unitlength}{\unitlength * \real{\svgscale}}%
    \fi%
  \else%
    \setlength{\unitlength}{\svgwidth}%
  \fi%
  \global\let\svgwidth\undefined%
  \global\let\svgscale\undefined%
  \makeatother%
  \begin{picture}(1,0.66666667)%
    \lineheight{1}%
    \setlength\tabcolsep{0pt}%
    \put(0,0){\includegraphics[width=\unitlength,page=1]{retina_50.pdf}}%
    \put(0.16022728,0.04954063){\makebox(0,0)[t]{\lineheight{1.25}\smash{\begin{tabular}[t]{c}0.0\end{tabular}}}}%
    \put(0,0){\includegraphics[width=\unitlength,page=2]{retina_50.pdf}}%
    \put(0.30113637,0.04954063){\makebox(0,0)[t]{\lineheight{1.25}\smash{\begin{tabular}[t]{c}0.2\end{tabular}}}}%
    \put(0,0){\includegraphics[width=\unitlength,page=3]{retina_50.pdf}}%
    \put(0.44204546,0.04954063){\makebox(0,0)[t]{\lineheight{1.25}\smash{\begin{tabular}[t]{c}0.4\end{tabular}}}}%
    \put(0,0){\includegraphics[width=\unitlength,page=4]{retina_50.pdf}}%
    \put(0.58295455,0.04954063){\makebox(0,0)[t]{\lineheight{1.25}\smash{\begin{tabular}[t]{c}0.6\end{tabular}}}}%
    \put(0,0){\includegraphics[width=\unitlength,page=5]{retina_50.pdf}}%
    \put(0.7238636,0.04954063){\makebox(0,0)[t]{\lineheight{1.25}\smash{\begin{tabular}[t]{c}0.8\end{tabular}}}}%
    \put(0,0){\includegraphics[width=\unitlength,page=6]{retina_50.pdf}}%
    \put(0.86477273,0.04954063){\makebox(0,0)[t]{\lineheight{1.25}\smash{\begin{tabular}[t]{c}1.0\end{tabular}}}}%
    \put(0.51249999,0.01787836){\makebox(0,0)[t]{\lineheight{1.25}\smash{\begin{tabular}[t]{c}\footnotesize Recall\end{tabular}}}}%
    \put(0,0){\includegraphics[width=\unitlength,page=7]{retina_50.pdf}}%
    \put(0.1087963,0.09741762){\makebox(0,0)[rt]{\lineheight{1.25}\smash{\begin{tabular}[t]{r}0.0\end{tabular}}}}%
    \put(0,0){\includegraphics[width=\unitlength,page=8]{retina_50.pdf}}%
    \put(0.1087963,0.18893277){\makebox(0,0)[rt]{\lineheight{1.25}\smash{\begin{tabular}[t]{r}0.2\end{tabular}}}}%
    \put(0,0){\includegraphics[width=\unitlength,page=9]{retina_50.pdf}}%
    \put(0.1087963,0.28044792){\makebox(0,0)[rt]{\lineheight{1.25}\smash{\begin{tabular}[t]{r}0.4\end{tabular}}}}%
    \put(0,0){\includegraphics[width=\unitlength,page=10]{retina_50.pdf}}%
    \put(0.1087963,0.37196309){\makebox(0,0)[rt]{\lineheight{1.25}\smash{\begin{tabular}[t]{r}0.6\end{tabular}}}}%
    \put(0,0){\includegraphics[width=\unitlength,page=11]{retina_50.pdf}}%
    \put(0.1087963,0.46347823){\makebox(0,0)[rt]{\lineheight{1.25}\smash{\begin{tabular}[t]{r}0.8\end{tabular}}}}%
    \put(0,0){\includegraphics[width=\unitlength,page=12]{retina_50.pdf}}%
    \put(0.1087963,0.55499339){\makebox(0,0)[rt]{\lineheight{1.25}\smash{\begin{tabular}[t]{r}1.0\end{tabular}}}}%
    \put(0.05791015,0.335){\rotatebox{90}{\makebox(0,0)[t]{\lineheight{1.25}\smash{\begin{tabular}[t]{c}\footnotesize Precision\end{tabular}}}}}%
    \put(0,0){\includegraphics[width=\unitlength,page=13]{retina_50.pdf}}%
    \put(0.51249999,0.60055556){\makebox(0,0)[t]{\lineheight{1.25}\smash{\begin{tabular}[t]{c}IoU = 0.50 | mAP = 38.85\%\end{tabular}}}}%
    \put(0,0){\includegraphics[width=\unitlength,page=14]{retina_50.pdf}}%
    \put(0.45177227,0.54814309){\makebox(0,0)[lt]{\lineheight{1.25}\smash{\begin{tabular}[t]{l}\footnotesize Land | AP = 43.14\%\end{tabular}}}}%
    \put(0,0){\includegraphics[width=\unitlength,page=15]{retina_50.pdf}}%
    \put(0.45177227,0.51342086){\makebox(0,0)[lt]{\lineheight{1.25}\smash{\begin{tabular}[t]{l}\footnotesize Mudflat | AP = 29.55\%\end{tabular}}}}%
    \put(0,0){\includegraphics[width=\unitlength,page=16]{retina_50.pdf}}%
    \put(0.45177227,0.47869864){\makebox(0,0)[lt]{\lineheight{1.25}\smash{\begin{tabular}[t]{l}\footnotesize Water | AP = 43.85\%\end{tabular}}}}%
  \end{picture}%
\endgroup%

%% file: figures/retina_75.pdf_tex
\begingroup%
  \makeatletter%
  \providecommand\color[2][]{%
    \errmessage{(Inkscape) Color is used for the text in Inkscape, but the package 'color.sty' is not loaded}%
    \renewcommand\color[2][]{}%
  }%
  \providecommand\transparent[1]{%
    \errmessage{(Inkscape) Transparency is used (non-zero) for the text in Inkscape, but the package 'transparent.sty' is not loaded}%
    \renewcommand\transparent[1]{}%
  }%
  \providecommand\rotatebox[2]{#2}%
  \newcommand*\fsize{\dimexpr\f@size pt\relax}%
  \newcommand*\lineheight[1]{\fontsize{\fsize}{#1\fsize}\selectfont}%
  \ifx\svgwidth\undefined%
    \setlength{\unitlength}{432bp}%
    \ifx\svgscale\undefined%
      \relax%
    \else%
      \setlength{\unitlength}{\unitlength * \real{\svgscale}}%
    \fi%
  \else%
    \setlength{\unitlength}{\svgwidth}%
  \fi%
  \global\let\svgwidth\undefined%
  \global\let\svgscale\undefined%
  \makeatother%
  \begin{picture}(1,0.66666667)%
    \lineheight{1}%
    \setlength\tabcolsep{0pt}%
    \put(0,0){\includegraphics[width=\unitlength,page=1]{retina_75.pdf}}%
    \put(0.16022728,0.04954063){\makebox(0,0)[t]{\lineheight{1.25}\smash{\begin{tabular}[t]{c}0.0\end{tabular}}}}%
    \put(0,0){\includegraphics[width=\unitlength,page=2]{retina_75.pdf}}%
    \put(0.30113637,0.04954063){\makebox(0,0)[t]{\lineheight{1.25}\smash{\begin{tabular}[t]{c}0.2\end{tabular}}}}%
    \put(0,0){\includegraphics[width=\unitlength,page=3]{retina_75.pdf}}%
    \put(0.44204546,0.04954063){\makebox(0,0)[t]{\lineheight{1.25}\smash{\begin{tabular}[t]{c}0.4\end{tabular}}}}%
    \put(0,0){\includegraphics[width=\unitlength,page=4]{retina_75.pdf}}%
    \put(0.58295455,0.04954063){\makebox(0,0)[t]{\lineheight{1.25}\smash{\begin{tabular}[t]{c}0.6\end{tabular}}}}%
    \put(0,0){\includegraphics[width=\unitlength,page=5]{retina_75.pdf}}%
    \put(0.7238636,0.04954063){\makebox(0,0)[t]{\lineheight{1.25}\smash{\begin{tabular}[t]{c}0.8\end{tabular}}}}%
    \put(0,0){\includegraphics[width=\unitlength,page=6]{retina_75.pdf}}%
    \put(0.86477273,0.04954063){\makebox(0,0)[t]{\lineheight{1.25}\smash{\begin{tabular}[t]{c}1.0\end{tabular}}}}%
    \put(0.51249999,0.01787836){\makebox(0,0)[t]{\lineheight{1.25}\smash{\begin{tabular}[t]{c}\footnotesize Recall\end{tabular}}}}%
    \put(0,0){\includegraphics[width=\unitlength,page=7]{retina_75.pdf}}%
    \put(0.1087963,0.09741762){\makebox(0,0)[rt]{\lineheight{1.25}\smash{\begin{tabular}[t]{r}0.0\end{tabular}}}}%
    \put(0,0){\includegraphics[width=\unitlength,page=8]{retina_75.pdf}}%
    \put(0.1087963,0.18893277){\makebox(0,0)[rt]{\lineheight{1.25}\smash{\begin{tabular}[t]{r}0.2\end{tabular}}}}%
    \put(0,0){\includegraphics[width=\unitlength,page=9]{retina_75.pdf}}%
    \put(0.1087963,0.28044792){\makebox(0,0)[rt]{\lineheight{1.25}\smash{\begin{tabular}[t]{r}0.4\end{tabular}}}}%
    \put(0,0){\includegraphics[width=\unitlength,page=10]{retina_75.pdf}}%
    \put(0.1087963,0.37196309){\makebox(0,0)[rt]{\lineheight{1.25}\smash{\begin{tabular}[t]{r}0.6\end{tabular}}}}%
    \put(0,0){\includegraphics[width=\unitlength,page=11]{retina_75.pdf}}%
    \put(0.1087963,0.46347823){\makebox(0,0)[rt]{\lineheight{1.25}\smash{\begin{tabular}[t]{r}0.8\end{tabular}}}}%
    \put(0,0){\includegraphics[width=\unitlength,page=12]{retina_75.pdf}}%
    \put(0.1087963,0.55499339){\makebox(0,0)[rt]{\lineheight{1.25}\smash{\begin{tabular}[t]{r}1.0\end{tabular}}}}%
    \put(0.05791015,0.335){\rotatebox{90}{\makebox(0,0)[t]{\lineheight{1.25}\smash{\begin{tabular}[t]{c}\footnotesize Precision\end{tabular}}}}}%
    \put(0,0){\includegraphics[width=\unitlength,page=13]{retina_75.pdf}}%
    \put(0.51249999,0.60055556){\makebox(0,0)[t]{\lineheight{1.25}\smash{\begin{tabular}[t]{c}IoU = 0.75 | mAP = 14.10\%\end{tabular}}}}%
    \put(0,0){\includegraphics[width=\unitlength,page=14]{retina_75.pdf}}%
    \put(0.45177227,0.54814309){\makebox(0,0)[lt]{\lineheight{1.25}\smash{\begin{tabular}[t]{l}\footnotesize Land | AP = 14.03\%\end{tabular}}}}%
    \put(0,0){\includegraphics[width=\unitlength,page=15]{retina_75.pdf}}%
    \put(0.45177227,0.51342086){\makebox(0,0)[lt]{\lineheight{1.25}\smash{\begin{tabular}[t]{l}\footnotesize Mudflat | AP = 10.41\%\end{tabular}}}}%
    \put(0,0){\includegraphics[width=\unitlength,page=16]{retina_75.pdf}}%
    \put(0.45177227,0.47869864){\makebox(0,0)[lt]{\lineheight{1.25}\smash{\begin{tabular}[t]{l}\footnotesize Water | AP = 17.85\%\end{tabular}}}}%
  \end{picture}%
\endgroup%

%% file: figures/faster_50.pdf_tex
\begingroup%
  \makeatletter%
  \providecommand\color[2][]{%
    \errmessage{(Inkscape) Color is used for the text in Inkscape, but the package 'color.sty' is not loaded}%
    \renewcommand\color[2][]{}%
  }%
  \providecommand\transparent[1]{%
    \errmessage{(Inkscape) Transparency is used (non-zero) for the text in Inkscape, but the package 'transparent.sty' is not loaded}%
    \renewcommand\transparent[1]{}%
  }%
  \providecommand\rotatebox[2]{#2}%
  \newcommand*\fsize{\dimexpr\f@size pt\relax}%
  \newcommand*\lineheight[1]{\fontsize{\fsize}{#1\fsize}\selectfont}%
  \ifx\svgwidth\undefined%
    \setlength{\unitlength}{432bp}%
    \ifx\svgscale\undefined%
      \relax%
    \else%
      \setlength{\unitlength}{\unitlength * \real{\svgscale}}%
    \fi%
  \else%
    \setlength{\unitlength}{\svgwidth}%
  \fi%
  \global\let\svgwidth\undefined%
  \global\let\svgscale\undefined%
  \makeatother%
  \begin{picture}(1,0.66666667)%
    \lineheight{1}%
    \setlength\tabcolsep{0pt}%
    \put(0,0){\includegraphics[width=\unitlength,page=1]{faster_50.pdf}}%
    \put(0.16022728,0.04954063){\makebox(0,0)[t]{\lineheight{1.25}\smash{\begin{tabular}[t]{c}0.0\end{tabular}}}}%
    \put(0,0){\includegraphics[width=\unitlength,page=2]{faster_50.pdf}}%
    \put(0.30113637,0.04954063){\makebox(0,0)[t]{\lineheight{1.25}\smash{\begin{tabular}[t]{c}0.2\end{tabular}}}}%
    \put(0,0){\includegraphics[width=\unitlength,page=3]{faster_50.pdf}}%
    \put(0.44204546,0.04954063){\makebox(0,0)[t]{\lineheight{1.25}\smash{\begin{tabular}[t]{c}0.4\end{tabular}}}}%
    \put(0,0){\includegraphics[width=\unitlength,page=4]{faster_50.pdf}}%
    \put(0.58295455,0.04954063){\makebox(0,0)[t]{\lineheight{1.25}\smash{\begin{tabular}[t]{c}0.6\end{tabular}}}}%
    \put(0,0){\includegraphics[width=\unitlength,page=5]{faster_50.pdf}}%
    \put(0.7238636,0.04954063){\makebox(0,0)[t]{\lineheight{1.25}\smash{\begin{tabular}[t]{c}0.8\end{tabular}}}}%
    \put(0,0){\includegraphics[width=\unitlength,page=6]{faster_50.pdf}}%
    \put(0.86477273,0.04954063){\makebox(0,0)[t]{\lineheight{1.25}\smash{\begin{tabular}[t]{c}1.0\end{tabular}}}}%
    \put(0.51249999,0.01787836){\makebox(0,0)[t]{\lineheight{1.25}\smash{\begin{tabular}[t]{c}\footnotesize Recall\end{tabular}}}}%
    \put(0,0){\includegraphics[width=\unitlength,page=7]{faster_50.pdf}}%
    \put(0.1087963,0.09741762){\makebox(0,0)[rt]{\lineheight{1.25}\smash{\begin{tabular}[t]{r}0.0\end{tabular}}}}%
    \put(0,0){\includegraphics[width=\unitlength,page=8]{faster_50.pdf}}%
    \put(0.1087963,0.18893277){\makebox(0,0)[rt]{\lineheight{1.25}\smash{\begin{tabular}[t]{r}0.2\end{tabular}}}}%
    \put(0,0){\includegraphics[width=\unitlength,page=9]{faster_50.pdf}}%
    \put(0.1087963,0.28044792){\makebox(0,0)[rt]{\lineheight{1.25}\smash{\begin{tabular}[t]{r}0.4\end{tabular}}}}%
    \put(0,0){\includegraphics[width=\unitlength,page=10]{faster_50.pdf}}%
    \put(0.1087963,0.37196309){\makebox(0,0)[rt]{\lineheight{1.25}\smash{\begin{tabular}[t]{r}0.6\end{tabular}}}}%
    \put(0,0){\includegraphics[width=\unitlength,page=11]{faster_50.pdf}}%
    \put(0.1087963,0.46347823){\makebox(0,0)[rt]{\lineheight{1.25}\smash{\begin{tabular}[t]{r}0.8\end{tabular}}}}%
    \put(0,0){\includegraphics[width=\unitlength,page=12]{faster_50.pdf}}%
    \put(0.1087963,0.55499339){\makebox(0,0)[rt]{\lineheight{1.25}\smash{\begin{tabular}[t]{r}1.0\end{tabular}}}}%
    \put(0.05791015,0.335){\rotatebox{90}{\makebox(0,0)[t]{\lineheight{1.25}\smash{\begin{tabular}[t]{c}\footnotesize Precision\end{tabular}}}}}%
    \put(0,0){\includegraphics[width=\unitlength,page=13]{faster_50.pdf}}%
    \put(0.51249999,0.60055556){\makebox(0,0)[t]{\lineheight{1.25}\smash{\begin{tabular}[t]{c}IoU = 0.50 | mAP = 37.21\%\end{tabular}}}}%
    \put(0,0){\includegraphics[width=\unitlength,page=14]{faster_50.pdf}}%
    \put(0.45177227,0.54814309){\makebox(0,0)[lt]{\lineheight{1.25}\smash{\begin{tabular}[t]{l}\footnotesize Land | AP = 39.31\%\end{tabular}}}}%
    \put(0,0){\includegraphics[width=\unitlength,page=15]{faster_50.pdf}}%
    \put(0.45177227,0.51342086){\makebox(0,0)[lt]{\lineheight{1.25}\smash{\begin{tabular}[t]{l}\footnotesize Mudflat | AP = 31.17\%\end{tabular}}}}%
    \put(0,0){\includegraphics[width=\unitlength,page=16]{faster_50.pdf}}%
    \put(0.45177227,0.47869864){\makebox(0,0)[lt]{\lineheight{1.25}\smash{\begin{tabular}[t]{l}\footnotesize Water | AP = 41.14\%\end{tabular}}}}%
  \end{picture}%
\endgroup%

%% file: figures/faster_75.pdf_tex
\begingroup%
  \makeatletter%
  \providecommand\color[2][]{%
    \errmessage{(Inkscape) Color is used for the text in Inkscape, but the package 'color.sty' is not loaded}%
    \renewcommand\color[2][]{}%
  }%
  \providecommand\transparent[1]{%
    \errmessage{(Inkscape) Transparency is used (non-zero) for the text in Inkscape, but the package 'transparent.sty' is not loaded}%
    \renewcommand\transparent[1]{}%
  }%
  \providecommand\rotatebox[2]{#2}%
  \newcommand*\fsize{\dimexpr\f@size pt\relax}%
  \newcommand*\lineheight[1]{\fontsize{\fsize}{#1\fsize}\selectfont}%
  \ifx\svgwidth\undefined%
    \setlength{\unitlength}{432bp}%
    \ifx\svgscale\undefined%
      \relax%
    \else%
      \setlength{\unitlength}{\unitlength * \real{\svgscale}}%
    \fi%
  \else%
    \setlength{\unitlength}{\svgwidth}%
  \fi%
  \global\let\svgwidth\undefined%
  \global\let\svgscale\undefined%
  \makeatother%
  \begin{picture}(1,0.66666667)%
    \lineheight{1}%
    \setlength\tabcolsep{0pt}%
    \put(0,0){\includegraphics[width=\unitlength,page=1]{faster_75.pdf}}%
    \put(0.16022728,0.04954063){\makebox(0,0)[t]{\lineheight{1.25}\smash{\begin{tabular}[t]{c}0.0\end{tabular}}}}%
    \put(0,0){\includegraphics[width=\unitlength,page=2]{faster_75.pdf}}%
    \put(0.30113637,0.04954063){\makebox(0,0)[t]{\lineheight{1.25}\smash{\begin{tabular}[t]{c}0.2\end{tabular}}}}%
    \put(0,0){\includegraphics[width=\unitlength,page=3]{faster_75.pdf}}%
    \put(0.44204546,0.04954063){\makebox(0,0)[t]{\lineheight{1.25}\smash{\begin{tabular}[t]{c}0.4\end{tabular}}}}%
    \put(0,0){\includegraphics[width=\unitlength,page=4]{faster_75.pdf}}%
    \put(0.58295455,0.04954063){\makebox(0,0)[t]{\lineheight{1.25}\smash{\begin{tabular}[t]{c}0.6\end{tabular}}}}%
    \put(0,0){\includegraphics[width=\unitlength,page=5]{faster_75.pdf}}%
    \put(0.7238636,0.04954063){\makebox(0,0)[t]{\lineheight{1.25}\smash{\begin{tabular}[t]{c}0.8\end{tabular}}}}%
    \put(0,0){\includegraphics[width=\unitlength,page=6]{faster_75.pdf}}%
    \put(0.86477273,0.04954063){\makebox(0,0)[t]{\lineheight{1.25}\smash{\begin{tabular}[t]{c}1.0\end{tabular}}}}%
    \put(0.51249999,0.01787836){\makebox(0,0)[t]{\lineheight{1.25}\smash{\begin{tabular}[t]{c}\footnotesize Recall\end{tabular}}}}%
    \put(0,0){\includegraphics[width=\unitlength,page=7]{faster_75.pdf}}%
    \put(0.1087963,0.09741762){\makebox(0,0)[rt]{\lineheight{1.25}\smash{\begin{tabular}[t]{r}0.0\end{tabular}}}}%
    \put(0,0){\includegraphics[width=\unitlength,page=8]{faster_75.pdf}}%
    \put(0.1087963,0.18893277){\makebox(0,0)[rt]{\lineheight{1.25}\smash{\begin{tabular}[t]{r}0.2\end{tabular}}}}%
    \put(0,0){\includegraphics[width=\unitlength,page=9]{faster_75.pdf}}%
    \put(0.1087963,0.28044792){\makebox(0,0)[rt]{\lineheight{1.25}\smash{\begin{tabular}[t]{r}0.4\end{tabular}}}}%
    \put(0,0){\includegraphics[width=\unitlength,page=10]{faster_75.pdf}}%
    \put(0.1087963,0.37196309){\makebox(0,0)[rt]{\lineheight{1.25}\smash{\begin{tabular}[t]{r}0.6\end{tabular}}}}%
    \put(0,0){\includegraphics[width=\unitlength,page=11]{faster_75.pdf}}%
    \put(0.1087963,0.46347823){\makebox(0,0)[rt]{\lineheight{1.25}\smash{\begin{tabular}[t]{r}0.8\end{tabular}}}}%
    \put(0,0){\includegraphics[width=\unitlength,page=12]{faster_75.pdf}}%
    \put(0.1087963,0.55499339){\makebox(0,0)[rt]{\lineheight{1.25}\smash{\begin{tabular}[t]{r}1.0\end{tabular}}}}%
    \put(0.05791015,0.335){\rotatebox{90}{\makebox(0,0)[t]{\lineheight{1.25}\smash{\begin{tabular}[t]{c}\footnotesize Precision\end{tabular}}}}}%
    \put(0,0){\includegraphics[width=\unitlength,page=13]{faster_75.pdf}}%
    \put(0.51249999,0.60055556){\makebox(0,0)[t]{\lineheight{1.25}\smash{\begin{tabular}[t]{c}IoU = 0.75 | mAP = 14.00\%\end{tabular}}}}%
    \put(0,0){\includegraphics[width=\unitlength,page=14]{faster_75.pdf}}%
    \put(0.46650028,0.54814309){\makebox(0,0)[lt]{\lineheight{1.25}\smash{\begin{tabular}[t]{l}\footnotesize Land | AP = 15.35\%\end{tabular}}}}%
    \put(0,0){\includegraphics[width=\unitlength,page=15]{faster_75.pdf}}%
    \put(0.46650028,0.51342086){\makebox(0,0)[lt]{\lineheight{1.25}\smash{\begin{tabular}[t]{l}\footnotesize Mudflat | AP = 9.26\%\end{tabular}}}}%
    \put(0,0){\includegraphics[width=\unitlength,page=16]{faster_75.pdf}}%
    \put(0.46650028,0.47869864){\makebox(0,0)[lt]{\lineheight{1.25}\smash{\begin{tabular}[t]{l}\footnotesize Water | AP = 17.38\%\end{tabular}}}}%
  \end{picture}%
\endgroup%